# Evaluation The Efficiency Of Cuckoo Optimization Algorithm

[1]Elham Shadkam and [2]Mehdi Bijari

[1, 2] Department of Industrial and Systems Engineering, Isfahan University of Technology, Isfahan, Iran

## ABSTRACT

*In this paper a new evolutionary algorithm, for continuous nonlinear optimization problems, is surveyed. This method is inspired by the life of a bird, called Cuckoo.*

*The Cuckoo Optimization Algorithm (COA) is evaluated by using the Rastrigin function. The problem is a non-linear continuous function which is used for evaluating optimization algorithms. The efficiency of the COA has been studied by obtaining optimal solution of various dimensions Rastrigin function in this paper. The mentioned function also was solved by FA and ABC algorithms. Comparing the results shows the COA has better performance than other algorithms.*

*Application of algorithm to test function has proven its capability to deal with difficult optimization problems.*

## 1. INTRODUCTION

Many problems are continuous in the real world and finding the global solutions is difficult. Although the development in computer technologies is increasing the speed of computations, this often is not adequate, particularly if the size of the problem's instance is large. Applying exact algorithm on such problems necessitate their linearization. Heuristic methods have been tackling the problems within reasonable computational time. Heuristic methods give an approximate solution.

The late studies of the researchers have led to development of the algorithms which have been based on the natural phenomenon. Several Meta-Heuristic algorithms have been developed during recent decades. For solving the combined optimization problems, The Meta-Heuristic methods have been efficient in finding the solution [1, 2, 3, 4].

Many Meta-Heuristic algorithms have been presented basing the nature of which the Particle Swarm Optimization (PSO) [5], Artificial Bee Colony (ABC) [6], Firefly Algorithm (FA) [7], Bee Colony Optimization (BCO) [8] and Ants Colony Optimization (ACO) [9] could be mentioned. The ABC algorithm [6, 10] is based on the mining manner of the bee colony for solving the continuous optimization problems. The FA [7] is one of the Meta-Heuristic algorithms inspired by the swarm intelligence for continuous optimization problems.
Some presented methods apply clustering decision space idea. Like the novel approach which used for scheduling problems [11]. To solve this problem the proposed heuristic based on finding approximate jobs' positions in a sequence and then applying the induced knowledge to find more precise jobs' situation.

In this paper, we evaluate the ABC algorithm and FA to show the efficiency of COA algorithm and solve some continuous optimization functions and answers will be compared.

Cuckoo Optimization Algorithm is based on the life of a bird called cuckoo. The basis of this novel optimization algorithm is Specific breeding and egg laying of this bird. Adult cuckoos and eggs used in this modeling. Adult cuckoos lay eggs in other birds' habitat. Those eggs grow and become a mature cuckoo if are not fiends and not removed by host birds. The immigration of groups of cuckoos and environmental specifications hopefully lead them to converge and reach the best place for reproduction and breeding. The global maximum of objective functions is in this best palace.

The rest of the paper is organized as follows: in part 2, we've presented COA algorithm; in part 3, the evaluation of COA, ABC and FA for resolving the continuous optimization problems has been studied and we've considered the results. In part 4, conclusions are presented.

## 2. CUCKOO'S LIFESTYLE

Cuckoo optimization was developed by Yang and Deb in 2009 that inspired from the nature [12]. Cuckoo Optimization Algorithm was developed by Rajabioun in 2011 [13]. Cuckoo Optimization Algorithm (COA) is really a new continuous over all aware search based on the life of a bird called cuckoo. Similar other meta heuristic, COA begins with an primary population, a group of cuckoos. These cuckoos lay some eggs in the habitat of other host birds. A random group of potential solutions is generated that represent the habitat in COA. In the fitness function will be evaluated the parameters of the candidate components. The steps to search the optimum solution are given as follows. First, the algorithm begins with a primary population of birds and they have some eggs to lie in some host bird's habitats. Some of these eggs grow up and become adult birds which are more like to the host bird's eggs and other eggs are discovered by host birds and are removed. The more profit is gained in that area, the more eggs that remain and hatch in the place. So the place in which more eggs remain will be the term that COA is going to optimize. Each cuckoo has a "distance" to the best habitat.

It is essential that the values of the problem be changed as an array, called "habitat" for solving a problem with COA. A habitat is an array of $1*N_{var}$, indicating the current living place of the cuckoo in the instance of $N_{var}$ dimensional case,. It is defined as follows:

*habitat=[$x_1, x_2, ..., x_{Nvar}$ ]*       (1)

The suitability of a habitat is yielded by assessment of profit function $f_b$ at a habitat of ($x_1, x_2, ..., x_{Nvar}$), where:

*profit=$f_b$(habitat)=$f_b$($x_1, x_2, ..., x_{Nvar}$)*       (2)

To begin the COA, a primary habitat matrix of size $N_{pop}*N_{var}$ is introduced. For each of these primary cuckoo habitats, some randomly produced number of eggs is supposed. the cuckoos lay eggs within a maximum interval from their position, this domain is called Egg Laying Radius.

Egg Laying Radius (ELR) is given as follows:

$$ELR = \alpha \times \frac{Number\ of\ current\ cuckoo's\ eggs}{Total\ number\ of\ eggs}(var_{High} - var_{Low})$$       (3)

handles the maximum value of ELR and is an integer value. $var_{hi}$ and $var_{low}$ are the upper and the lower bound for variables respectively. After each egg laying, P% of all the eggs (usually 10%) that less similar to the eggs of the host bird are discovered and thrown out of the nest. Therefore, the eggs that profit function value is smaller remove.

Remained young cuckoos are grown in the host nests. The adult cuckoos live for a while in their groups and habitats when they grow up. Then they migrate to better places where the eggs have higher survival chances. Cuckoo groups are shaped in all different place of the environment. For migration, the group with the best situation is selected as the goal point for all other cuckoos. it is a problem to find out to which group each cuckoo belongs.

Grouping of birds is done by K-means clustering method to solve this problem. Group's relative optimization of habitat is calculated based on the average objective function of the group. Other groups migrate to the place with the greatest average profit. The cuckoos do not migrate the entire path in travel to the goal point; they just pass part of the distance. This subject is clearly shown in Figure1:

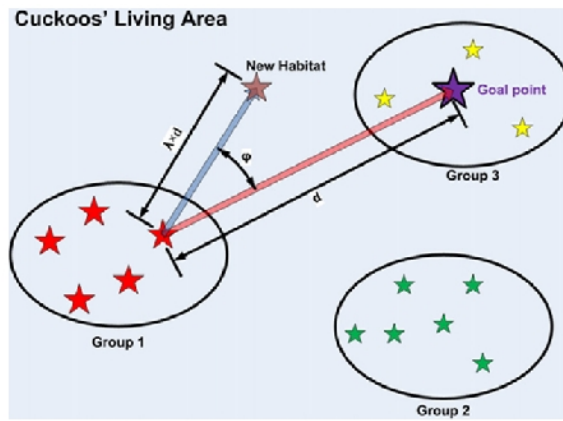

Fig 1. Immigration of a sample cuckoo toward goal habitat [13]

Each bird just traverse % of the whole path to the goal place and has a deviation of radians. and help the birds search a larger area. is a random number between /6 and – /6 and is a number between 0 and 1. Each cuckoo lays some eggs, once all the cuckoos have traveled to the goal place and the entire new place are specified.

An ELR is determined for each cuckoo based on the number of eggs. The maximum number of the bird can live in a place is finite ($N_{max}$). After some iteration, all of the cuckoos achieve an optimized place with the maximum likeness of the eggs to the eggs of the host bird. This position the number of the eggs which removed in it will be minimal and will have the greatest objective function. Takes algorithm to its end if convergence of more than 95% of all the cuckoos towards a single point. Relationship (2), migration function in COA, is:

$$X_{next\ habitate} = X_{current\ habitate} + F \times (X_{Goalpoint} - X_{current\ habitate}) \quad (4)$$

The cuckoo optimization algorithm (COA) is summarized as follows [14] and Fig. 2 shows a flowchart of the proposed algorithm.

1. Preparation cuckoo habitats with some arbitrary solution on the objective function;
2. Assignment some eggs to each cuckoo;
3. Determination ELR for each cuckoo;
4. Allow cuckoo to lay eggs inside their equivalent ELR;

5. Removed those eggs that are discovered by host birds;
6. Allow eggs hatch and chicks grow up;
7. Assess the habitat of each grown cuckoo;
8. Remove cuckoo live in worst habitats and confine maximum number of cuckoo in environment;
9. Cluster cuckoos and detect best place and select goal point;
10. Allow new cuckoo population travel toward goal point;
11. If stop condition is satisfied, then stop. If not, go to 2.

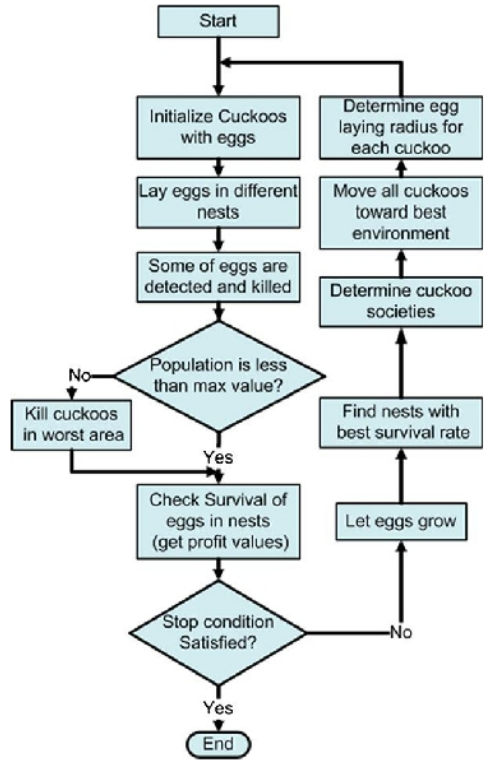

Fig 2. Flowchart of cuckoo optimization algorithm [13]

## 3. EVALUATION AND RESULTS

For evaluation and finding out efficiency of the COA, the Rastrigin function is studied [15]. This function has several maximum and minimum points which have caused it to be utilize as a test function for assessment of the Meta-Heuristic Algorithms. So, the Rastrigin function is used for comparison of the evaluation and utility of COA algorithm.

$$f = 10n + \sum_{i=1}^{n}(x_i^2 - 10\cos(2\pi x_i)), \quad f(0,0,\ldots,0) = 0, \quad -5 \leq x_i \leq 5 \quad (5)$$

Rastrigin function is one of the difficult test functions and has plenty of local minimal, even in 3-dimensional case. Fig. 3 shows the 3-dimensional Rastrigin function. The Rastrigin function is a really challenging optimization problem, as it is appear even in 3-dimensional case.

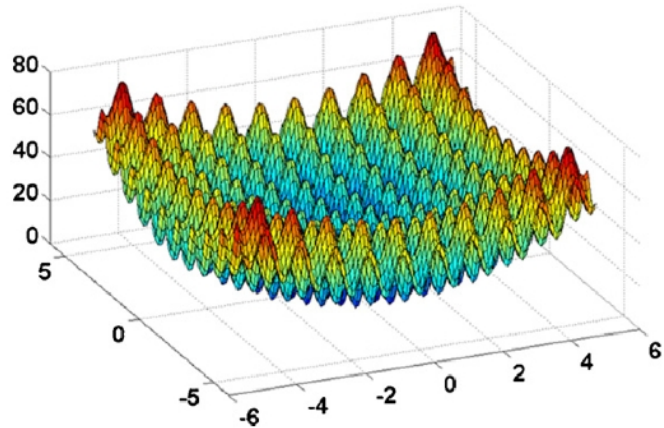

Fig 3. 3D plot of Rastrigin function

The Meta-Heuristic algorithms are very sensitive for their parameters and the setting of the parameters can affect their efficiency. The parameters settings cause more reliability and flexibility of the algorithm. So, settings of the parameters are one of the crucial factors in gaining the optimized solution in all optimization problems. Table 1 shows the selected parameters for COA algorithm.

Table 1. Parameters settings for COA algorithm

| Max number of eggs | Min number of eggs | Number of initial population | Higher limitation of variable | Lower limitation of variable |
|---|---|---|---|---|
| 4 | 2 | 5 | 30 | -30 |
| Population variance that cuts the optimization | Control parameter of egg laying (Radius Coeff) | Max Number of cuckoos | Lambda variable (Motion Coeff) | Number of clusters |
| 1e-13 | 5 | 10 | 9 | 2 |

The results of Table 2 demonstrate that using the COA, ABC and FA algorithm makes getting the optimized solution possible. The number of initial population is set 50 For ABC and FA algorithms and the number of iterations is set 100 for both algorithms. As can be seen, the COA algorithm reaches the optimum value. The results of the ABC and FA algorithms are derived from [16].

Table 2. Comparison of the Results

| Dimensional of Function | Range of search Space | ABC | FA | COA |
|---|---|---|---|---|
| 2 | ±30 | 0.0059 | 0.1287 | 0 |
| 3 | ±30 | 0.0136 | 0.7516 | 0 |

To show the efficiency of COA, the convergence diagram is used. As shown in figure 4, the function of the algorithms in convergence toward the optimal solution in appropriate number of repetitions.

International Journal on Computational Sciences & Applications (IJCSA) Vol.4, No.2, April 2014

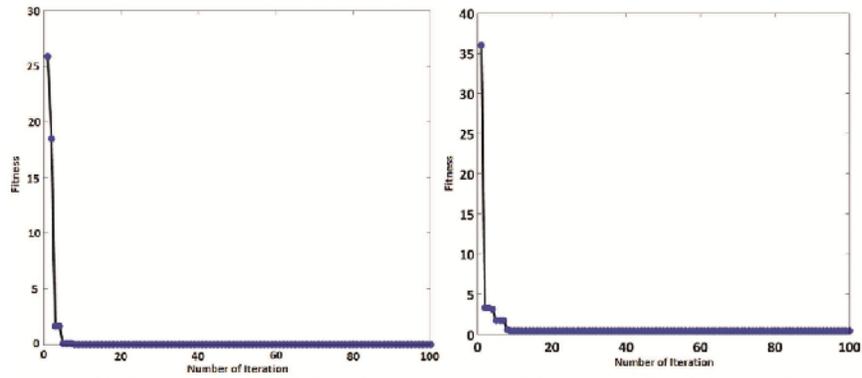

ABC for Solving the Two Dimensional Rastrigin [12]    ABC for Solving the Three Dimensional Rastrigin[12]

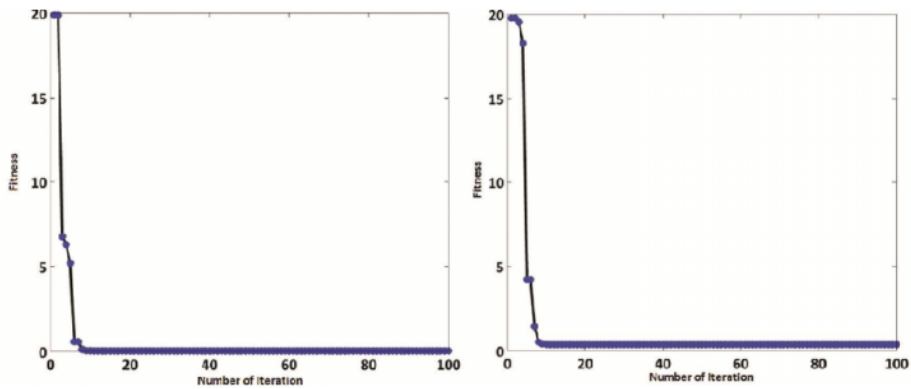

FA for Solving the Two Dimensional Rastrigin [16]  FA for Solving the Three Dimensional Rastrigin [16]

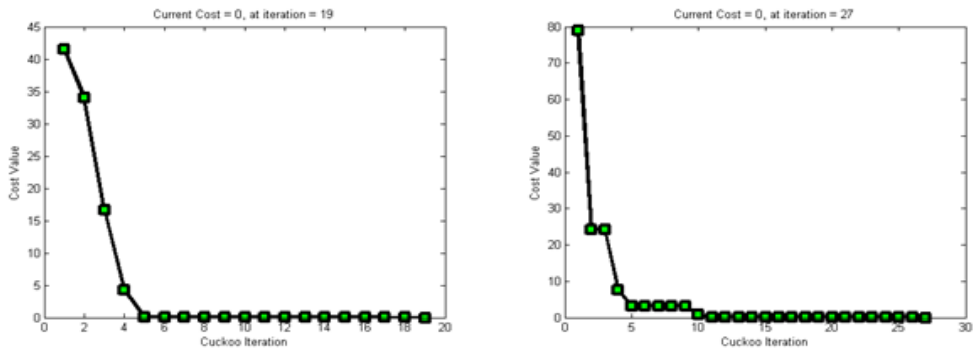

COA for Solving the Two Dimensional Rastrigin    COA for Solving the Three Dimensional Rastrigin

Fig 4. The Convergence Diagram of COA, ABC and FA for Solving the Two and Three Dimensional Rastrigin Function

Table 3 shows the number of the repetition of the execution of COA, ABC and FA for two dimensional Rastrigin function. The results of table 3 and fig 5, show the fact that COA algorithm is more efficient in finding the global optimized points.

Table 3. The result of comparison for two dimensional Rastrigin function

| Algorithm | Iteration | | | | |
|---|---|---|---|---|---|
| | 20 | 40 | 60 | 80 | 100 |
| ABC | 1.0372 | 1.1610 | 1.0056 | 0.0148 | 0.0059 |
| FA | 1.5273 | 1.3604 | 1.1312 | 1.0228 | 0.1287 |
| COA | 4.26E-14 | 0 | 0 | 0 | 0 |

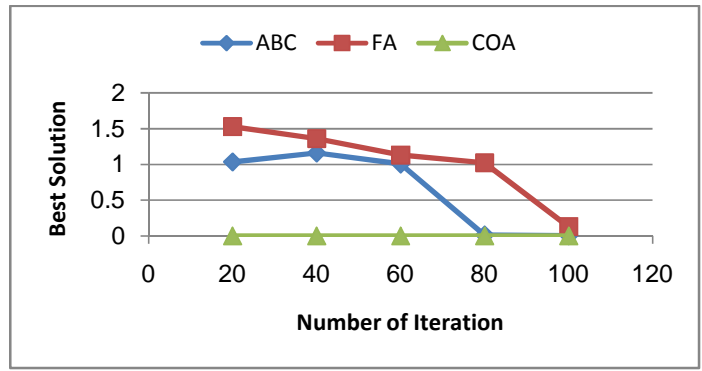

Fig 5. The Comparison Diagram for two dimensional Rastrigin function

Table 4 shows the repetition number of execution of COA, ABC and FA for three dimensional Rastrigin function. As it is seen, COA algorithm is more able in finding the global optimized points and is more efficient in solving the continuous optimization functions of large dimensions.

Table 3. The result of comparison for three dimensional Rastrigin function

| Algorithm | Iteration | | | | |
|---|---|---|---|---|---|
| | 20 | 40 | 60 | 80 | 100 |
| ABC | 1.6723 | 1.1688 | 1.6165 | 0.0826 | 0.0136 |
| FA | 2.2512 | 2.1298 | 1.6228 | 1.1743 | 0.7516 |
| COA | 7.01E-12 | 0 | 0 | 0 | 0 |

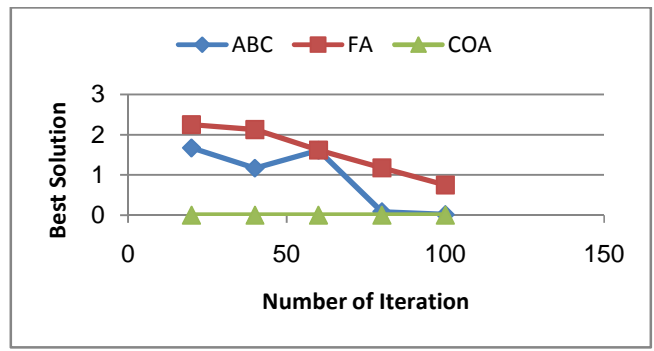

Fig 6. The Comparison between tree algorithms for three dimensional Rastrigin function

COA algorithm is able to maintain the equilibrium of the local and global search of the problem in spite of the increase in the dimensions of it in an optimized way, as it is seen from table 4.

COA algorithm reaches the optimal solution (zero value) for different dimension in Finite number of iterations.

Table 4. Number of Iteration to reach the optimal solution

| Dimensional of Function | 2 | 3 | 5 | 10 | 50 | 100 | 1000 | 10000 |
|---|---|---|---|---|---|---|---|---|
| Number of Iteration | 25 | 28 | 30 | 32 | 35 | 36 | 38 | 39 |

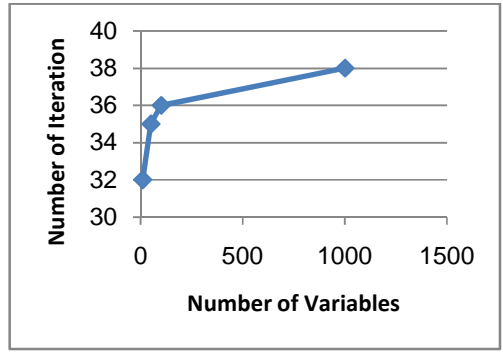

Fig 7. Number of Iteration to reach the optimal solution

Table 5 shows variation and mean of solutions obtained from the COA algorithm. Results indicate the robustness of COA algorithm is compared to the dimension of the problem (number of iteration=20).

Table 5. variation and mean of solutions obtained from the COA algorithm

| Dimension of the problem | 2 | 3 | 5 | 10 | 50 | 100 | 1000 | 10000 |
|---|---|---|---|---|---|---|---|---|
| Mean of solution | 3.70E-07 | 1.5485E-05 | 3.81027E-05 | 4.38E-06 | 3.97E-05 | 5.66E-05 | 0.01405 | 0.06668 |
| Variation of solution | 8.05424E-07 | 7.40086E-10 | 2.49416E-08 | 5.63182E-11 | 5.12347E-09 | 4.82102E-09 | 0.000283 | 0.00032 65 |

## 4. CONCLUSION

In this paper, a novel optimization algorithm was verified which was based on lifestyle of a bird called Cuckoo. Egg laying and breeding are Special characteristics of cuckoos and had been the basic motivation for development of this algorithm. The COA algorithm was evaluated on benchmark cost functions. We have studied the COA, ABC and FA for solving the continuous optimization problems of big area and the answers near to the optimized one. To evaluate the efficiency of the algorithm, two types of comparison from answer reliability and accuracy in convergence points have been studied. The comparison of COA with ABC and FA, showed the superiority of COA in fast convergence and global optima achievement. In this test, other methods have not found the global minimal but COA found it and COA has converged faster in less iterations. In the test function (low dimensional Rastrigin function) methods have found the near global minima but COA has converged faster in less iterations. But in test function with high dimensional, ABC and FA could not converge to even a near value of global optimal. But COA had found global minimum in just 35 iterations. COA algorithm is stronger from answering point and more appropriate in solving such problems and is also when it is convergent; it is faster than other algorithm.

# REFERENCES

[1] F.S. Gharehchopogh, I. Maleki, S.R. Khaze, "A New Optimization Method for Dynamic Travelling Salesman Problem with Hybrid Ant Colony Optimization Algorithm and Particle Swarm Optimization", International Journal of Advanced Research in Computer Engineering & Technology (IJARCET), Vol. 2, Issue 2, pp. 352-358, February 2013.

[2] F.S. Gharehchopogh, I.Maleki, S.R.Khaze, "A New Approach in Dynamic Traveling Salesman Solution: A Hybrid of Ant Colony Optimization and Descending Gradients", International Journal of Managing Public Sector Information and Communication Technologies (IJMPICT), Vol. 3, No. 2, pp. 1-9, December 2012.

[3] I. Maleki, S.R. Khaze, F.S. Gharehchopogh, "A New Solution for Dynamic Travelling Salesman Problem with Hybrid Ant Colony Optimization Algorithm and Chaos Theory", International Journal of Advanced Research in Computer Science (IJARCS), Vol. 3, No. 7, pp. 39-44, Nov-Dec 2012.

[4] F.S. Gharehchopogh, I. Maleki, M. Farahmandian, "New Approach for Solving Dynamic Traveling Salesman Problem with Hybrid Genetic Algorithms and Ant Colony Optimization", International Journal of Computer Applications (IJCA), Vol. 53, No.1, pp. 39-44, September 2012.

[5] J. Kennedy, R.C. Eberhart, "Particle Swarm Optimization", In Proceedings of the IEEE International Conference on Neural Networks, pp. 1942-1948, 1995.

[6] D. Karaboga, "An idea based on honeybee swarm for numerical optimization", Technical Report TR06, Erciyes University, Engineering Faculty, Computer Engineering Department, 2005.

[7] X.S. Yang, "Nature-Inspired Meta-heuristic Algorithms", Luniver Press, 2008.

[8] P. Lucic, D. Teodorovic, "Bee system: Modeling Combinatorial Optimization Transportation Engineering Problems by Swarm Intelligence", Preprints of the TRISTAN IV Triennial Symposium on Transportation Analysis, Sao Miguel, Azores Islands, pp. 441-445, 2001.

[9] M. Dorigo, V. Maniezzo, A. Colorni, "Ant system: optimization by a colony of cooperating agents", IEEE Trans. on Systems, Man and Cybernetics, Part B, Vol.26, No.1, pp.29-41, 1996.

[10] P. Shunmugapriya, S. Kanmani, "Artificial Bee Colony Approach for Optimizing Feature Selection", International Journal of Computer Science Issues, Vol. 9, Issue 3, No. 3, pp. 432-438, May 2012.

[11] S. Poursheikh Ali, M. Bijari, " Minimizing Maximum Earliness and Tardiness on a Single Machine using a Novel Heuristic Approach", Proceedings of the International Conference on Industrial Engineering and Operations Management, Istanbul, Turkey, July 3 – 6, 2012.

[12] X. S. Yang and S. Deb, "Cuckoo search via Lévy Flights," In: World Congress on Nature & Biologically Inspired Computing (NaBIC2009). IEEE Publications, pp. 210–214, 2009.

[13] R. Rajabioun, "Cuckoo Optimization Algorithm," In: Applied Soft Computing journal, vol. 11, pp. 5508 - 5518, 2011.

[14] M. Arezki Mellal, s. Adjerid, E. Williams and D. Benazzouz, "Optimal replacement policy for bsolete components using cuckoo optimization algorithm based-approach: Dependability context," Journal of Scientific & Industrial research, Vol. 71, pp. 715-721, 2012.

[15] M. Molga, C. Smutnicki, "Test functions for optimization needs", 2005.

[16] R. Khaze, I. Maleki, S. Hojjatkhah and A. Bagherinia, "evaluation the efficiency of artificial bee colony and the firefly algorithm in solving the continues optimization problem", International Journal on Computational Sciences & Applications (IJCSA), Vol.3, No.4, pp.23-35, 2013.

**Authors**

**Elham shadkam** is a PhD student in the Department of Industrial Engineering at Isfahan University of Technology in Isfahan, Iran. Her main research interests are modeling and simulation, Optimization, Multi objective decision making.

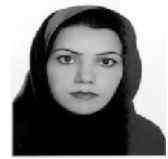

**Mehdi Bijari** received his BSc in Industrial Engineering 1987, Msc in system planning 1990, both from Isfahan University of Technology (IUT), and PhD in Industrial Engineering 2002, Sharif University of Technology. He has lectured in the Industrial Engineering Department at IUT from 1991. He is Associate professor in Industrial Engineering. His interested research areas are in the Stochastic production planning, Meta Heuristics Methods, Optimization and Artificial Intelligence. He has published more than 50 papers in journals and conferences.

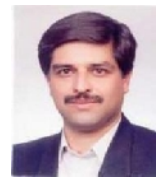